\documentclass[graybox]{svmult}

\usepackage{mathptmx}       % selects Times Roman as basic font
\usepackage{helvet}         % selects Helvetica as sans-serif font
\usepackage{courier}        % selects Courier as typewriter font
\usepackage{type1cm}        % activate if the above 3 fonts are
\usepackage{makeidx}         % allows index generation
\usepackage{graphicx}        % standard LaTeX graphics tool
\usepackage{multicol}        % used for the two-column index
\usepackage[bottom]{footmisc}% places footnotes at page bottom

\makeindex             % used for the subject index
\usepackage{geometry}
\usepackage{graphicx}
\usepackage{wrapfig}
\usepackage{listings}
\usepackage{color}
\usepackage{xcolor}
\usepackage{amssymb}
\usepackage{hyperref}

\newcommand*\samethanks[1][\value{footnote}]{\footnotemark[#1]}

\begin{document}

\title*{Learning to Run challenge: Synthesizing physiologically accurate motion using deep reinforcement learning}

\author{{\L}ukasz Kidzi\'nski, Sharada P. Mohanty\thanks{These authors contributed equally to this work}, Carmichael Ong\samethanks[1], Jennifer L. Hicks, Sean F. Carroll, Sergey Levine, Marcel Salath\'e, Scott L. Delp}
\date{\today}

\maketitle

\abstract{
%Make abstract a single paragraph?
Synthesizing physiologically-accurate human movement in a variety of conditions can help practitioners plan surgeries, design experiments, or prototype assistive devices in simulated environments, reducing time and costs and improving treatment outcomes. Because of the large and complex solution spaces of biomechanical models, current methods are constrained to specific movements and models, requiring careful design of a controller and hindering many possible applications.\\
We sought to discover if modern optimization methods efficiently explore these complex spaces. To do this, we posed the problem as a competition in which participants were tasked with developing a controller to enable a physiologically-based human model to navigate a complex obstacle course as quickly as possible, without using any experimental data. They were provided with a human musculoskeletal model and a physics-based simulation environment.\\
In this paper, we discuss the design of the competition, technical difficulties, results, and analysis of the top controllers. The challenge proved that deep reinforcement learning techniques, despite their high computational cost, can be successfully employed as an optimization method for synthesizing physiologically feasible motion in high-dimensional biomechanical systems.
}

\section{Overview of the competition}
Human movement results from the intricate coordination of muscles, tendons, joints, and other physiological elements. While children learn to walk, run, climb, and jump in their first years of life and most of us can navigate complex environments---like a crowded street or moving subway---without considerable active attention, developing controllers that can efficiently and robustly synthesize realistic human motions in a variety of environments remains a grand challenge for biomechanists, neuroscientists, and computer scientists. Current controllers are confined to a small set of pre-specified movements or driven by torques, rather than the complex muscle actuators found in humans (see Section \ref{ss:background}).

%The figure was created by Samuel Hamner, I believe ... he should be given credit % Ł: I emailed him and addded 'Image courtesy of Samuel Hamner'
\begin{figure}[ht!]
\centering
\includegraphics[width=0.8\linewidth]{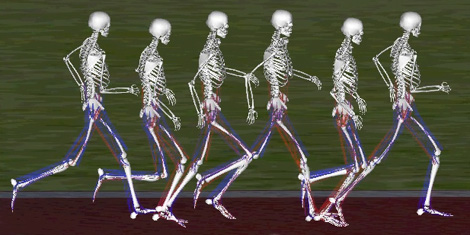}
\caption{Musculoskeletal simulation of human running in Stanford's OpenSim software. OpenSim was used to simulate the muscoloskeletal lower-body system used in the competition, and the competitors were tasked with learning controllers that could actuate the muscles in the presence of realistic delays to achieve rapid running gaits. Image courtesy of Samuel Hamner.}
\label{fig:running}
\end{figure}

In this competition, participants were tasked with developing a controller to enable a physiologically-based human model to navigate a complex obstacle course as quickly as possible. Participants were provided with a human musculoskeletal model and a physics-based simulation environment where they could synthesize physically and physiologically accurate motion (Figure \ref{fig:running}). Obstacles were divided into two groups: external and internal. External obstacles consisted of soft balls fixed to the ground to create uneven terrain, and internal obstacles included introducing weakness in the psoas muscle, a key muscle for swinging the leg forward during running. Controllers submitted by the participants were scored based on the distance the agents equipped with these controllers traveled through the obstacle course in a set amount of time. 
% The following sentence seems a little out of place. I also edited
To simulate the fact that humans typically move in a manner that minimizes the risk of joint injury, controllers were penalized for excessive use of ligament forces. We provided competitors with a set of training environments to help build robust controllers; competitors' scores were based on a final, unknown environment that used more external obstacles (10 balls instead of 3) in an unexpected configuration (see Section \ref{ss:humanoid-model}).

The competition was designed for participants to use reinforcement learning methods to create their controllers; however, participants were allowed to use other optimization frameworks. As the benchmark, we used state-of-the art reinforcement learning techniques: Trust Region Policy Optimization (TRPO) \cite{schulman2015trust} and Deep Deterministic Policy Gradients (DDPG) \cite{lillicrap2015continuous}. We included implementations of these reinforcement learning models in the ``Getting Started'' tutorial provided to competitors.

This competition fused biomechanics, computer science, and neuroscience to explore a grand challenge in human movement and motor control. The entire competition was built on free and open source software. Participants were required to tackle three major challenges in reinforcement learning: large dimensionality of the action space, delayed actuation, and robustness to variability of the environments. Controllers that can synthesize the movement of realistic human models can help optimize human performance (e.g., fine-tune technique for high jump or design a prosthetic to break paralympic records) and plan surgery and treatment for individuals with movement disorders (see Section \ref{ss:background}). In Section \ref{s:results}, we analyze accuracy of the top results from a biomechanical standpoint and discuss implications of the results and propose future directions. For a description of solutions from top participants please refer to \cite{jaskowski2018rltorunfast,kidzinski2018l2rsolutions}.

To the best of our knowledge, this was the largest reinforcement learning competition in terms of the number of participants and the most complex in terms of environment, to date. In Section \ref{s:organization} we share our insights from the process of designing the challenge and our solutions to problems encountered while administering the challenge.

%\subsection*{Keywords}
%biomechanics, musculoskeletal models, motor control, reinforcement learning, gait 

\section{Prior work}

We identify two groups of prior challenges related to this proposal. The first set includes challenges held within the biomechanics community, including the Dynamic Walking Challenge\footnote{\url{http://simtk-confluence.stanford.edu:8080/pages/viewpage.action?pageId=5113821}} (exploring mechanics of a very simple 2D walker) and the Grand Challenge Competition to Predict In Vivo Knee Loads\footnote{\url{https://simtk.org/projects/kneeloads}} (validation of musculoskeletal model estimates of muscle and joint contact forces in the knee). In the Dynamic Walking Challenge, the model used was highly simplified to represent the minimimum viable model to achieve bipedal gait without muscles. In the Grand Challenge, the focus was to predict knee loads given a prescribed motion rather than to generate novel motions.

The second class of prior challenges has been held in the reinforcement learning community. In the field of reinforcement learning, competitions have periodically been organized around standardized benchmark tasks\footnote{see, e.g., \url{http://www.rl-competition.org/}}. These tasks are typically designed to drive advancements in algorithm efficiency, exploration, and scalability. Many of the formal competitions, however, have focused on relatively smaller tasks, such as simulated helicopter control \cite{dimitrakakis2014reinforcement}, where the state and action dimensionality are low. More recently, vision-based reinforcement learning tasks, such as the Arcade Learning Environment (ALE) \cite{bellemare13arcade} have gained popularity. Although ALE was never organized as a formal contest, the Atari games in ALE have frequently been used as benchmark tasks in evaluating reinforcement algorithms with high-dimensional observations. However, these tasks do not test an algorithm's ability to learn to coordinate complex and realistic movements, as would be required for realistic running. The OpenAI gym benchmark tasks \cite{brockman2016openai} include a set of continuous control benchmarks based on the MuJoCo physics engine \cite{todorov2012mujoco}, and while these tasks do include bipedal running, the corresponding physical models use simple torque-driven frictionless joints, and successful policies for these benchmarks typically exhibit substantial visual artifacts and non-naturalistic gaits\footnote{see, e.g., \url{https://youtu.be/hx_bgoTF7bs}}. Furthermore, these tasks do not include many of the important phenomena involved in controlling musculoskeletal systems, such as delays.

There were three key features differentiating the ``Learning to Run'' challenge from other reinforcement learning competitions. First, in our competition, participants were tasked with building a robust controller for an unknown environment with external obstacles (balls fixed in the ground) and internal obstacles (reduced muscle strength), rather than a predefined course. Models experienced all available types of obstacles in the training environments, but competitors did not know how these obstacles would be positioned in the final test obstacle course. This novel aspect of our challenge forced participants to build more robust and generalizable solutions than for static environments such as those provided by OpenAI. Second, the dimensionality and complexity of the action space were much larger than in most popular reinforcement learning problems. It is comparable to the most complex MuJoCo physics OpenAI gym task \verb|Humanoid-V1| \cite{brockman2016openai} which had $17$ torque actuators, compared to $18$ actuators in this challenge. In contrast to many robotics competitions, the task in this challenge was to actuate muscles, which included delayed actuation and other physiological complexities, instead of controlling torques. This increased the complexity of the relationship between the control signal and torque generated. Furthermore, compared to torque actuators, more muscles are needed to fully actuate a model. Third, the cost of one iteration is larger, since precise simulations of muscles are computationally expensive. This constraint forces participants to build algorithms using fewer evaluations of the environment.

\section{Competition description}

\subsection{Background}\label{ss:background}

Understanding motor control is a grand challenge in biomechanics and neuroscience. One of the greatest hurdles is the complexity of the neuromuscular control system. Muscles are complex actuators whose forces are dependent on their length, velocity, and activation level, and these forces are then transmitted to the bones through a compliant tendon. Coordinating these musculotendon actuators to generate a robust motion is further complicated by delays in the biological system, including sensor delays, control signal delays, and muscle-tendon dynamics. Existing techniques allow us to estimate muscle activity from experimental data \cite{thelen2003cmc}, but solutions from these methods are insufficient to generate and predict new motions in a novel environment.

% the following statement is a bit bold. I might say "can help us solve ..." % Carmichael resolved by accepting
Recent advances in reinforcement learning, biomechanics, and neuroscience can help us solve this grand challenge. The biomechanics community has used single shooting methods to synthesize simulations of human movement driven by biologically inspired actuators. Early work directly solved for individual muscle excitations for a gait cycle of walking \cite{anderson2001dynamic} and for a maximum height jump \cite{anderson1999dynamic}. Recent work has focused on using controllers based on human reflexes to generate simulations of walking on level ground \cite{geyer2010muscle}. This framework has been extended to synthesize simulations of other gait patterns such as running \cite{wang2012optimizing}, loaded and inclined walking \cite{dorn2015predictive}, and turning and obstacle avoidance \cite{song2015neural}. Although these controllers were based on physiological reflexes, they needed substantial input from domain experts. Furthermore, use of these controllers has been limited to cyclic motions, such as walking and running, over static terrain.

Modern reinforcement learning techniques have been used recently to train more general controllers for locomotion. These techniques have the advantage that, compared to the gait controllers previously described, less user input is needed to hand tune the controllers, and they are more flexible to learning additional, novel tasks. For example, reinforcement learning has been used to train controllers for locomotion of complicated humanoid models \cite{lillicrap2015continuous,schulman2015trust}. Although these methods found solutions without domain specific knowledge, the resulting motions were not realistic. One possible reason for the lack of human-like motion is that these models did not use biologically accurate actuators.

Thus while designing the ``Learning to Run'' challenge, we conjectured that reinforcement learning methods would yield more realistic results with biologically accurate models and actuators. OpenSim is an open-source software environment which implements computational biomechanical models and allows muscle-driven simulations of these models \cite{delp2007opensim}. It is a flexible platform that can be easily incorporated into an optimization routine using reinforcement learning.

\subsection{OpenSim simulator}\label{ss:humanoid-model}

OpenSim is an open-source project that provides tools to model complex musculoskeletal systems in order to gain a better understanding of how movement is coordinated. OpenSim uses another open-source project, Simbody, as a dependency to perform the physics simulation. Users can employ either inverse methods, which estimate the forces needed to produce a given motion from data, or forward methods, which synthesize a motion from a set of controls. In this competition, we used OpenSim to 1) model the human musculoskeletal system and generate the corresponding equations of motion and 2) synthesize motions by integrating the equations of motion over time.

The human musculoskeletal model was based on a previous model \cite{delp1990model} and was simplified to decrease complexity, similarly to previous work \cite{ong2017walking}. The model was composed of 7 bodies. The pelvis, torso, and head were represented by a single body. Each leg had 3 bodies: an upper leg, a lower leg, and a foot. The model contained 9 degrees of freedom (dof): 3-dof between the pelvis and ground (i.e., two translation and one rotation), 1-dof hip joints, 1-dof knee joints, and 1-dof ankle joints. 

\begin{wrapfigure}[37]{r}{0.22\textwidth}
  \centering
  \includegraphics[width=\linewidth]{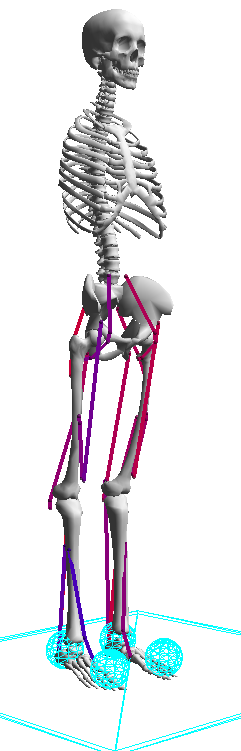}
  \caption{Musculoskeletal model in OpenSim used in this competition. Red/purple curves indicate muscles, while blue balls attached to feet model contact.}
  \label{fig:skeleton}
\end{wrapfigure}

The model included 18 musculotendon actuators \cite{thelen2003muscle}, with 9 on each leg, to represent the major lower limb muscle groups that drive walking (Figure \ref{fig:skeleton}). For each leg, these included the biarticular hamstrings, short head of the biceps femoris, gluteus maximus, iliopsoas, rectus femoris, vasti, gastrocnemius, soleus, and tibialis anterior. The force in these actuators mimicked biological muscle as the force depends on the length ($l$), velocity ($v$), and activation ($a$) level (i.e., the control signal to a muscle that is actively generating force, which can range between 0\% and 100\% activated) of the muscle. Biological muscle can produce force either actively, via a neural signal to the muscle to produce force, or passively, by being stretched past a certain length. The following equation shows how the total force was calculated, due to both active and passive force, in the each muscle ($F_{muscle}$),

\[F_{muscle} = F_{max-iso}(af_{active}(l)f_{velocity}(v) + f_{passive}(l)),\]

where $F_{max-iso}$ is the maximum isometric force of a muscle (i.e., a stronger muscle will have a larger value), $f_{active}$ and $f_{passive}$ are functions relating the active and passive force in a muscle to its current length, and $f_{velocity}$ is a function that scales the force a muscle produces as a function of its current velocity (e.g., a muscle can generate more force when lengthening than shortening). For a sense of scale, in this model, values of $F_{max-iso}$ ranged between 557 N and 9594 N. Force is transferred between the muscle and bone by tendons. Tendons are compliant structures that generate force when stretched beyond a certain length. Given the physical constraints between the tendon and muscle, a force equilibrium must be satisfied between them, governed by the relationship,

\[F_{tendon} = F_{muscle}\cos(\alpha),\]

where $\alpha$ is the pennation angle (i.e., the angle between the direction of the tendon and the muscle fibers).

Additionally, arbitrary amounts of force cannot be generated instantaneously due to various electrical, chemical, and mechanical delays in the biological system between an incoming electrical control signal and force generation. This was modeled using a first-order dynamic model between excitation (i.e., the neural signal as it reaches the muscle) and activation \cite{thelen2003muscle}.

The model also had components that represent ligaments and ground contact. Ligaments are biological structures that produce force when they are stretched past a certain length, protecting against excessively large joint angles. Ligaments were modeled at the hip, knee, and ankle joints as rotational springs with increasing stiffness as joint angle increases. These springs only engaged at larger flexion and extension angles. Ground contact was modeled using the Hunt-Crossley model \cite{hunt1975contact}, a compliant contact model. Two contact spheres were located at the heel and toes of each foot and generate forces depending on the depth and velocity of these spheres penetrating other contact geometry, including the ground, represented as a half-plane, and other obstacles, represented as other contact spheres. 

\subsection{Tasks and application scenarios}\label{ss:task}

In this competition, OpenSim and the model described in Section \ref{ss:humanoid-model} served as a black-box simulator of human movement. Competitors passed in excitations to each muscle, and OpenSim calculated and returned the state, which contained information about joint angles, joint velocities, body positions, body velocities, and distance to and size of the next obstacle. This occurred every 10 milliseconds during the simulation for 10 seconds (i.e., a total of 1000 decision time points).

%Competitors were tasked with building a real-time controller of a musculoskeletal model described in Section \ref{ss:humanoid-model} to navigate through an obstacle course as quickly as possible. 
% Do you mean the penalties on ligaments here? I don't think this has been defined.
%Apart from proxies for internal body sensors, observation included obstacle sensing.
% This was described in more detail previously. The extra detail seems to fit better here.
%At every decision time point,   We introduced internal and external obstacles. 
%Internal obstacles were varying muscle strengths across simulations. External obstacles were small balls fixed to the ground. Size and distance to the closest obstacle was included in the observation vector.

At every iteration the agent receives the current observed state vector $s \in \mathbb{R}^{41}$ consisting of the following:
\begin{itemize}
\item Rotations and angular velocities of the pelvis, hip, knee and ankle joints,
\item Positions and velocities of the pelvis, center of mass, head, torso, toes, and talus, 
\item Distance to the next obstacle (or 100 if it doesn't exist),
\item Radius and vertical location of the next obstacle.
\end{itemize}

Obstacles were small soft balls fixed to the ground. While it was possible to partly penetrate the ball, after stepping into the ball the repelling force was proportional to the volume of intersection of penetrating body and the ball. The first three balls were each positioned at a distance that was uniformly distributed between 1 and 5. Then, each subsequent obstacle was positioned at $u$ meters after the last one, where $u$ was uniformly distributed between 2 and 4. Each ball was fixed at $v$ meters vertically from the ground level, where $v$ was uniformly distributed between -0.25 and 0.25. Finally, the radius of each ball was $0.05 + v$, where $v$ was drawn from an exponential distribution with a mean of $0.05$. 

Based on the observation vector or internal states, current strength and distance to obstacles, participants' controllers were required to output a vector of current muscle excitations. These excitations were integrated over time to generate muscle activations (via a model of muscle's activation dynamics), which in turn generated movement (as a function of muscle moment arms and other muscle properties like strength and current length and lengthening velocity). Participants were evaluated by the distance they covered in a fixed amount of time. At every iteration the agent was expected to return a vector $v \in [0,1]^{18}$ of muscle excitations for the 18 muscles in the model.

Simulation environments were parametrized by: \verb|difficulty|, \verb|seed| and \verb|max_obstacles|. Difficulty corresponded to the number and density of obstacles. The \verb|seed| was a number which uniquely identifies pseudo-random generation of the obstacle positions in the environment and participants could use it in training to obtain a robust controller. The seed ranges between $0$ and $2^{63} - 1$. Both \verb|seed| and \verb|difficulty| of the final test environment were unknown to participants. Such a setup allowed us to give access to infinitely many training environments, as well as choose the final difficulty reactively, depending on users' performance leading up to the final competition round.

% first sentence of this paragraph is very strong, plus many will argue that it's not just the brain but also spinal cord, etc. either revise or remove % I changed it to "motor control system"
The controller modeled by participants was approximating functions of the human motor control system. It collected signals from physiological sensors and generated signals to excite muscles. Our objective was to construct the environment in such a way that its solutions could potentially help biomechanics and neuroscience researchers to better understand the mechanisms underlying human locomotion.

% i think this paragraph should be removed. doesn't seem like we should be trying to defend why it was feasible AFTER we ran the full challenge % removed
%To test if the problem is feasible, we ran a preliminary challenge with a simpler model\footnote{\url{https://www.crowdai.org/challenges/learning-how-to-walk}}. Results showed that modern deep reinforcement learning techniques can efficiently train a controller that quickly navigates through space.
%Compared to this preliminary competition, in the NIPS ``Learning to Run'' challenge we further developed the environment by including delayed action (naturally motivated by delays between muscle excitation and the generation of force) and obstacles. In order to tune the difficulty, we trained baseline controllers from the preliminary challenge (DDPG and TRPO) in the new setup. This allowed us to set the complexity at the right level: stimulating research, yet solvable.

\subsection{Baselines and code available}

% Maybe we shouldn't use the word "simple" here since our focus is that these models are more complex?  # done
%Is this referring to the previous walking challenge? If so, I think it could be fine to mention that we provided materials for a previous challenge in which competitors were tasked to get the same agent to walk? # done
Before running the NIPS competition, we organized a preliminary contest, with similar rules, to better understand feasibility of the deep reinforcement learning methods for the given task. We identified that existing deep learning techniques can be efficiently applied to the locomotion tasks with neuromusculoskeletal systems. Based on this experience, for the NIPS challenge, we used TRPO and DDPG as a baseline and we included implementation of a simple agent in the materials provided to the participants.

One of the objectives of the challenge was to bring together researchers from biomechanics, neuroscience and machine learning. We believe that this can only be achieved when entering the competition and building the most basic controller is seamless and takes seconds. To this end, we wrapped the sophisticated and complex OpenSim into a basic python environment with only two commands: \verb|reset(difficulty=0, seed=None)| and \verb|step(activations)|. The environment is freely available on GitHub\footnote{\url{https://github.com/stanfordnmbl/osim-rl}} and can be installed with 3 command lines on Windows, MacOS and Linux, using the Anaconda platform\footnote{\url{https://anaconda.org/}}. For more details regarding installation, refer to the Appendix.

\subsection{Metrics}
% Text in submissions section should all be full justified ... something funny is happening with the formatting
Submissions were evaluated automatically. Participants, after building the controller locally on their computers, were asked to interact with a remote environment. The objective of the challenge was to \textbf{navigate through the scene with obstacles to cover as much distance as possible in fixed time}. This objective was measured in meters from the origin on the $X$-axis the pelvis traveled during the simulation. To promote realistic solutions that avoid joint injury, we also introduced a penalty to the reward function for overusing ligaments.

We defined the objective function as

\[reward(T) = X(T) - \lambda \int_0^T \sqrt{L(t)}dt,\]

where $X(T)$ is the position of the pelvis at time $T$, $L(t)$ is the sum of squared forces generated by ligaments at time $t$ and $\lambda = 10^{-7}$ is a scaling factor. 
% So ligament forces were not actually penalized heavily? This contradicts statements earlier in the paper.
% What is learned by mentioning this "initial" mistake over just stating the current implementation of the objective function? % ligament forces are still in the objective function, but with a very small constant. It's actually hard to judge how much impact they have on the simulation (without running extra experiments) so I removed "almost negligible in the simulation"
The value of $\lambda$ was set very low due to an initial mistake in the system and it turns the impact of ligament forces smaller than we initially designed. The simulation was terminated either when the time reached $T=10s$ (equivalent to $1000$ simulation steps), or when the agent fell, which was defined as when the pelvis fell below $0.65m$.

In order to fairly compare the participants' controllers, the random seeds, determining muscle weakness and parameters of obstacles, were fixed for all the participants during grading.

\section{Organizational aspects}\label{s:organization}
\subsection{Protocol}

\begin{figure}[ht!]
  \centering
  \includegraphics[width=0.9\linewidth]{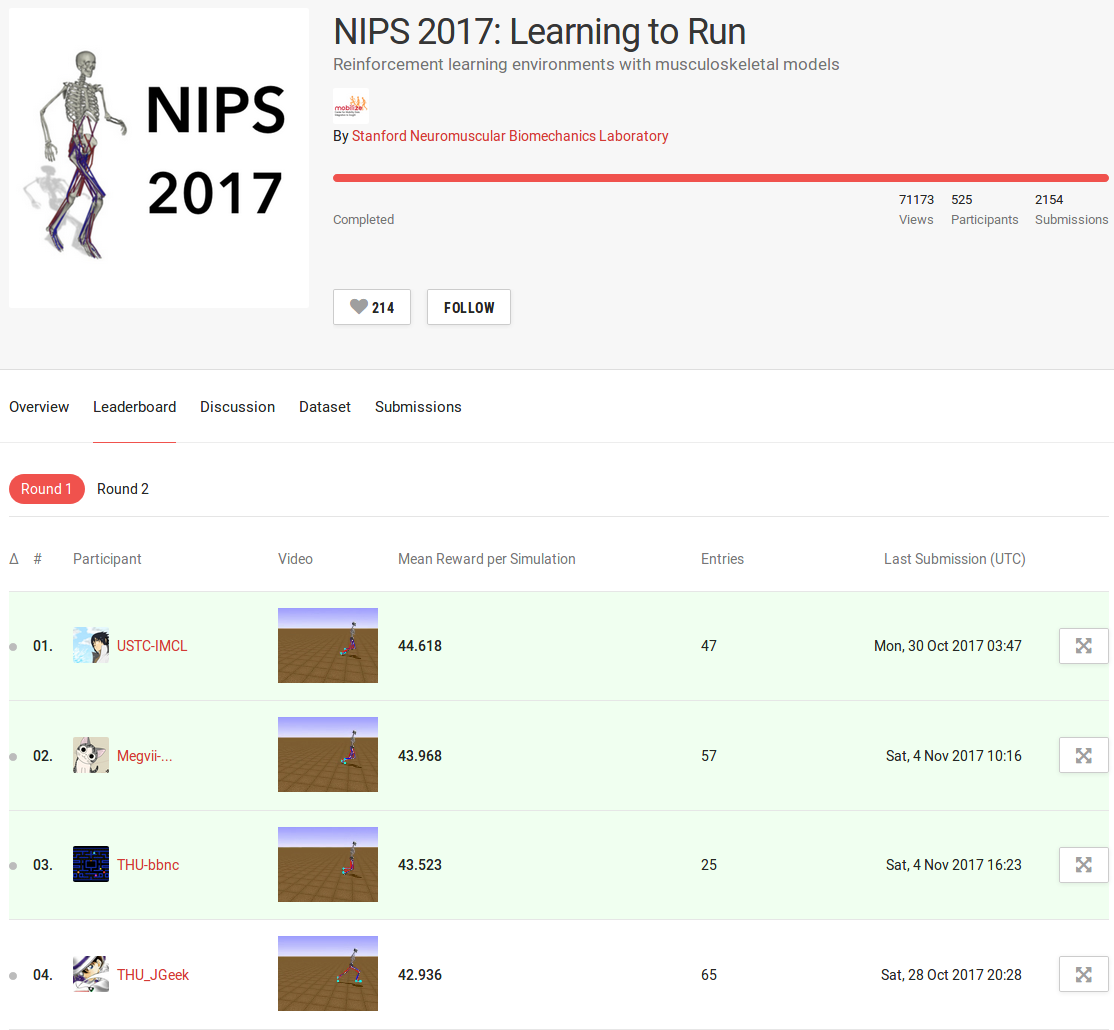}
  \caption{The leaderboard from the first round (Open Stage) of the ``Learning to Run'' challenge on the crowdAI.org platform. We conjecture that animated simulations contributed to engagement of participants.}
  \label{fig:crowdai}
\end{figure}

\begin{figure}[ht!]
  \centering
  \includegraphics[width=0.9\linewidth]{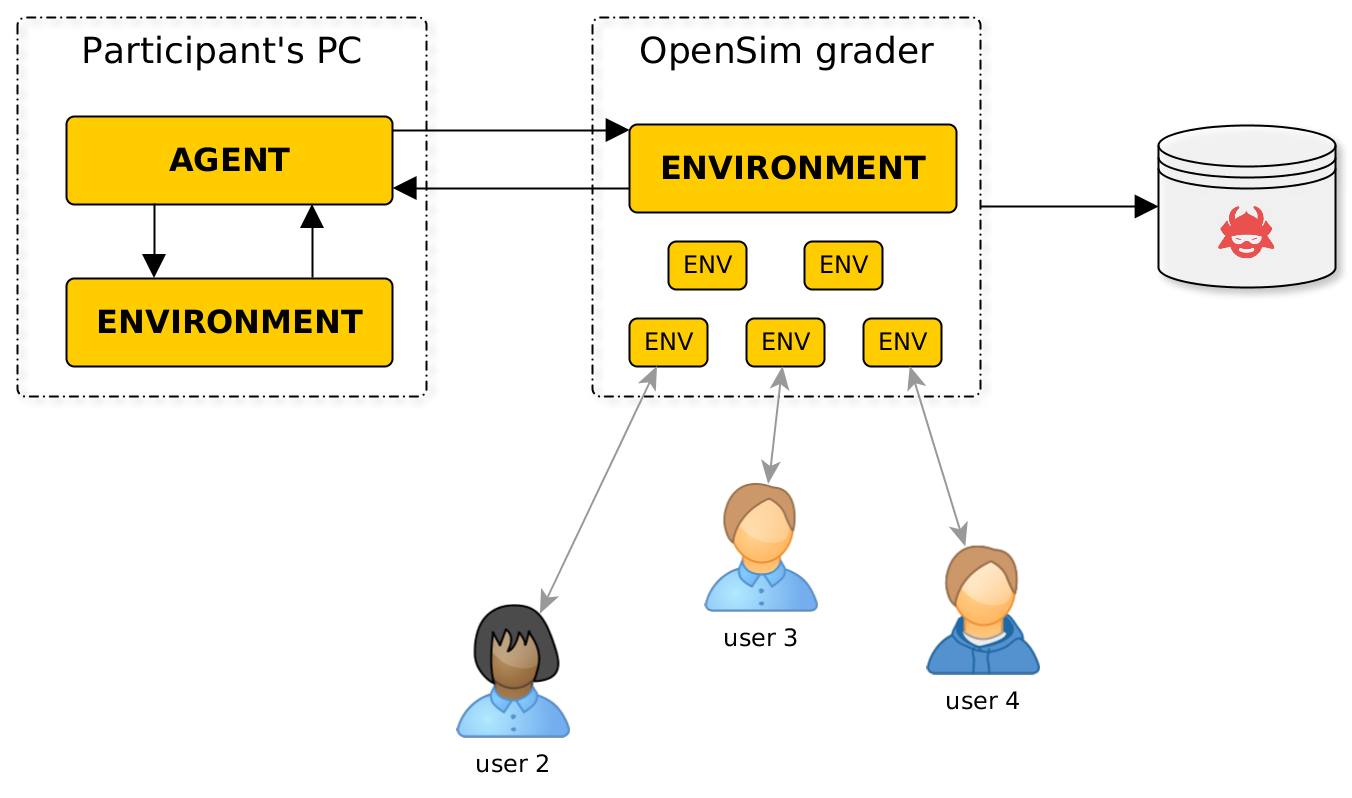}
  \caption{Schematic overview of the model training and submission process in the competition. Participants trained their agents on local machines, where they could run the environment freely. Once the agent was trained, they connected to the grader and, in an iterative process, they received the current state of the environment to which they responded with an action. After successful iteration until the end condition (10 seconds of a simulation or a fall of the agent), the final result was stored in the crowdAI database. Multiple users could connect to the grader simultaneously---each one to a separate environment. }
  \label{fig:flow}
\end{figure}

Participants were asked to register on the crowdAI.org\footnote{\url{http://crowdai.org/}} platform and download the ``Getting Started'' tutorial. The guide led participants through installation and examples of training baseline models (TRPO and DDPG). After training the model, participants connected to the grader and interacted with the remote environment, using a submission script that we provided (Figure \ref{fig:flow}). The remote environment iteratively sent the current observation and awaited response---the action of the participant in a given state. After that, the result was sent to the crowdAI.org platform and was listed on the leaderboard (as illustrated in Figure \ref{fig:crowdai}). Moreover, an animation corresponding to the best submission of a given user was displayed beside the score.

By interacting with the remote environment, participants could potentially explore it and tune their algorithms for the test environment. In order to prevent this exploration and overfitting, participants were allowed to send only five solutions per day. Moreover, the final score was calculated on a separate test environment to which users can submit only 3 solutions in total.

% it's unclear from above how the protocol allows for "penetration of the test environment"
% furthermore, i don't see why splitting the challenge into two key stages fixes this problem (reading further just makes it seem like changes were implemented in the play-off stage to address this issue but i don't know for sure????) % TODO
At the beginning of the challenge we did not know, how many participants to expect, or if the difficulty would be too low or too high. This motivated us to introduce two rounds:
\begin{enumerate}
\item The Open Stage was open for everyone and players were ranked by their result on the test environment. Every participant was allowed to submit 1 solution per day.
\item The Play-off Stage was open only for the competitors who earned at least 15 points in the Open Stage. Participants were allowed to submit only 3 solutions. Solutions were evaluated on a test environment different than the one in Open Stage.
\end{enumerate}
The Play-off Stage was open for one week after the Open Stage was finished. This setting allowed us to adjust the rules of the Play-off before it starts, while learning more about the problem and dynamics of the competition in the course of the Open Stage.

% the following paragraph comes out of nowhere. the previous paragarphs talk about preventing overfitting and this just talks about open source and engagement. could fit better somewhere else or taken out all together % removed for now
% We were using crowdAI.org platform with on-line submissions and a leaderboard. The platform has an established user base and it is open-source\footnote{\url{https://github.com/crowdAI/crowdai}}. On the leaderboard we presented short videos of moving models, in order to further engage participants to in the challenge.

We anticipated that the main source of cheating for locomotion tasks could be tracking of real or engineered data. To avoid this problem, we designed the competition such that competitors were scored on an unknown environment with obstacles, which means that a controller solely based on tracking is very unlikely to be successful.

To prevent overfitting as well as cheating, participants did not have access to the final test environment. Moreover, since participants were only interacting with a remote environment (as presented in Figure \ref{fig:flow}), they were not allowed to change parameters of the environment, such as gravity or obstacles. In fact, they were constrained to send only action vectors in $v \in [0,1]^{18}$ to the grader.

\subsection{Execution}\label{ss:execution}

In the Open Stage, participants interacted with the environment through a lightweight HTTP API included in the \verb|osim-rl| package. From a technical standpoint, in order to interact with a remote environment, they only needed to change the class from the local environment to HTTP API environment. The grader, on the remote host, was responsible for the life-cycle management of the environments. The cumulative rewards for each submission were added to the crowdAI leaderboard, along with visualization of the actual simulations. To judge submissions, 3 seeds for simulation environments were randomly chosen beforehand and were used to grade all submissions during this stage. % I tried to clean up the previous sentence for clarity. Please make sure I didn't introduce an error...

In the Play-off stage, participants packaged their agents into self-contained Docker containers. The containers would then interact with the grader using a lightweight redis API, simulating the process from the Open Stage. The grading infrastructure had a corresponding Docker image for the actual grading container. Grading a single submission involved instantiating the participant submitted Docker container, instantiating the internal grading container, mapping the relevant ports of the grading container and the submitted container, wrapping up both the containers in a separate isolated network, and then finally executing the pre-agreed grading script inside the participant submitted container.

\subsection{Problems and Solutions}

The major issues we encountered concerned the computational cost of simulations, over-fitting, and stochasticity of the results, i.e. high dependence of the random seed. Our solutions to each of these challenges are described below.
% the challenge of stochasticity was not clear to me

The Learning to Run environment was significantly slower than many visually similar environments such as the humanoid robot Mujoco-based simulations in OpenAI Gym\footnote{\url{https://github.com/stanfordnmbl/osim-rl/issues/78}}. This difference was due to the complex ground reaction model, muscle dynamics, and precision of simulation in OpenSim. Some of the participants modified the accuracy of OpenSim engine to trade off precision for execution speed\footnote{\url{https://github.com/ctmakro/stanford-osrl\#the-simulation-is-too-slow}}; even with these changes, the OpenSim-based Learning to Run environment continued to be expensive in terms of the actual simulation time. The computationally expensive nature of the problem required participants to find sample-efficient alternatives.

%i don't really follow the purpose of this paragraph. what's the main point? simulations were longer than expected? if so, this can be condenses quite a bit and the "story" aspect can be removed % I removed everything as it is not that important for other challenges
% In the Open Stage, the submissions were graded for a total of 3 simulations, and the participants were initially provided a time limit of 20 minutes for all the three simulations per submission. Because of the lack of previous benchmarks, the time limit was chosen based on the performance of the baseline models trained by the organizers, which had a mean cumulative reward of $\sim 21$. Midway through the competition, there were multiple complaints of timeouts during the submissions, and on a closer look we noticed that the average simulation times were much higher when the agents trained by the participants had mean cumulative rewards above 30, mostly because faster speed of running, leading to increased frequency of the contact force calculations between the musculoskeletal model and the ground. The time limit was adjusted to 90 minutes per submission, and continued to stay the same through the end of Open Stage. 

Another concern was the possibility of overfitting, since the random seed was fixed, and the number of simulations required for a fair evaluation of performance of submitted models. These issues were especially important in determining a clear winner during the Play-off stage. To address these issues, we based the design of the Play-off stage on Docker containers, as described in \ref{ss:execution}.

The design based on Docker containers has two main advantages for determining the top submissions. First, we could run an arbitrary number of simulations until we got performance scores for the top agents which were statistically significant. Given the observed variability of results in the Open Stage, we chose 10 simulations for the Play-off Stage and it proved to be sufficient for determining the winner. See Section \ref{ss:leaderboard} for details. Second, this setting prevents overfiting, since users do not have access to the test environment, while it allows us to use exactly the same environment (i.e., the same random seed) for every submission.

The main disadvantage of this design is the increased difficulty of submitting results, since it requires familiarity with the Docker ecosystem. For this reason, we decided to use this design only in the Play-off stage. This could potentially discourage participation. However, we conjectured that top participants who qualified to the Play-off stage will be willing to invest more time in preparing the submission, for the sake of fair and more deterministic evaluation. All top 10 participants from the Open Stage submitted their solutions to the Play-off stage.

%When participants have the objective to maximize their scores on the leaderboard based on just 3 simulations, there is a chance that they might tune their models to overfit the environments with the 3 specific seeds that were used by the grader for evaluation. This problem was partially solved by the Docker based grading approach of the Play-off Stage where participants had never interacted with the said 10 simulations that were used to evaluate their models. We also decided to release the simulation scores for all their submitted models only at the end of the challenge, to ensure participants have no feedback signal when choosing their final models, other than their own metric of generalizable performance of their models. 
% i don't fully understand what is going on here, but i can vaguely interpret what this means. this needs to be rewritten or cleaned up % removed
%Finally, some participants explored the environment for certain random seeds, so that they new positions of the obstacles in advance in future runs. They had consciously conditioned their models to those particular obstacle settings to perform well on the leaderboard. This problem concerned only the Open Stage where it was possible to interact with the test environment. The Docker based approach in the Play-off Stage ensured that the models submitted by participants had never been exposed to the final environments used for evaluation. %or that the participants did not get any feedback at all from the final environments when they were selecting and tuning their final models. 

\subsection{Submissions}\label{ss:leaderboard}

% \begin{figure}[ht!]
% \centering
% \includegraphics[width=\linewidth]{images/histogram.png}
% \caption{Distribution of the scores for all the submitted solutions. The dotted grey line represents the score (20.083) of the baseline submission.}
% \label{fig:score_distribution}
% \end{figure}

\begin{figure}[ht!]
\centering
\includegraphics[width=0.9\linewidth]{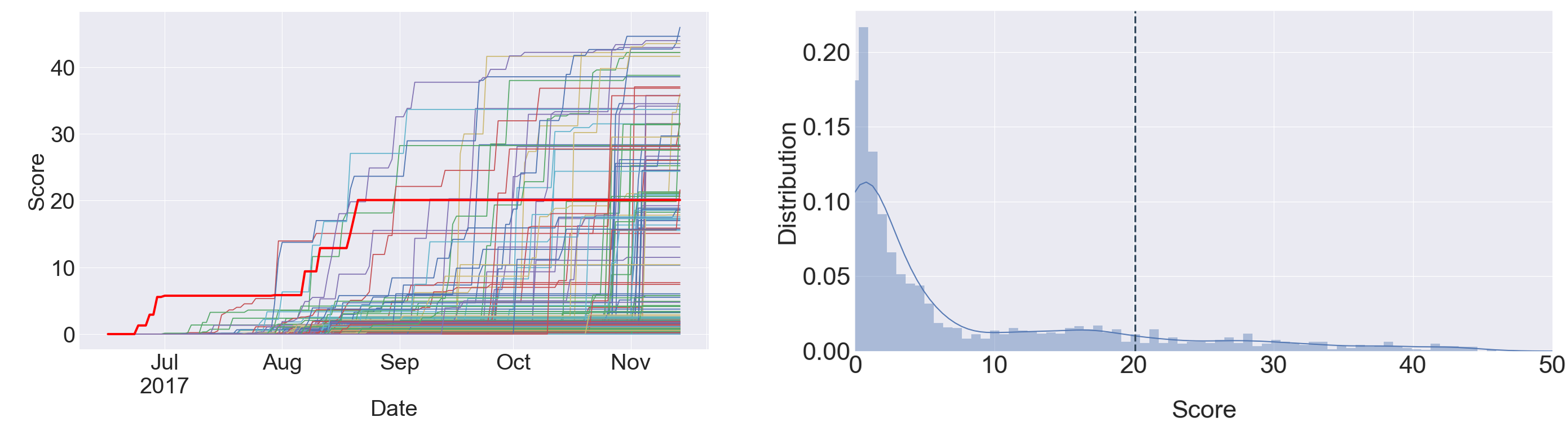}
\caption{Left: Progression of scores during the challenge. Each curve represents the maximum score of a single participant on the leaderboard at any point in time. The bold red line represents the baseline submission. 
Right: The final distribution of the scores for all the submitted solutions. The dotted gray line represents the score (20.083) of the baseline submission.}
\label{fig:score_progression}
\end{figure}

The competition was held between June 16th 2017 and November 13th 2017. It attracted $442$ teams with $2154$ submissions. The average number of submission was $4.37$ per team, with scores ranging between $-0.81$ and $45.97$ in both stages combined. In Figure~\ref{fig:score_progression} we present the progression of submissions over time.

% The y-axis labels of the left and right panel use different font sizes. It would also be good to match the size of each of these figures

\begin{figure}[ht!]
\centering
\includegraphics[width=\linewidth]{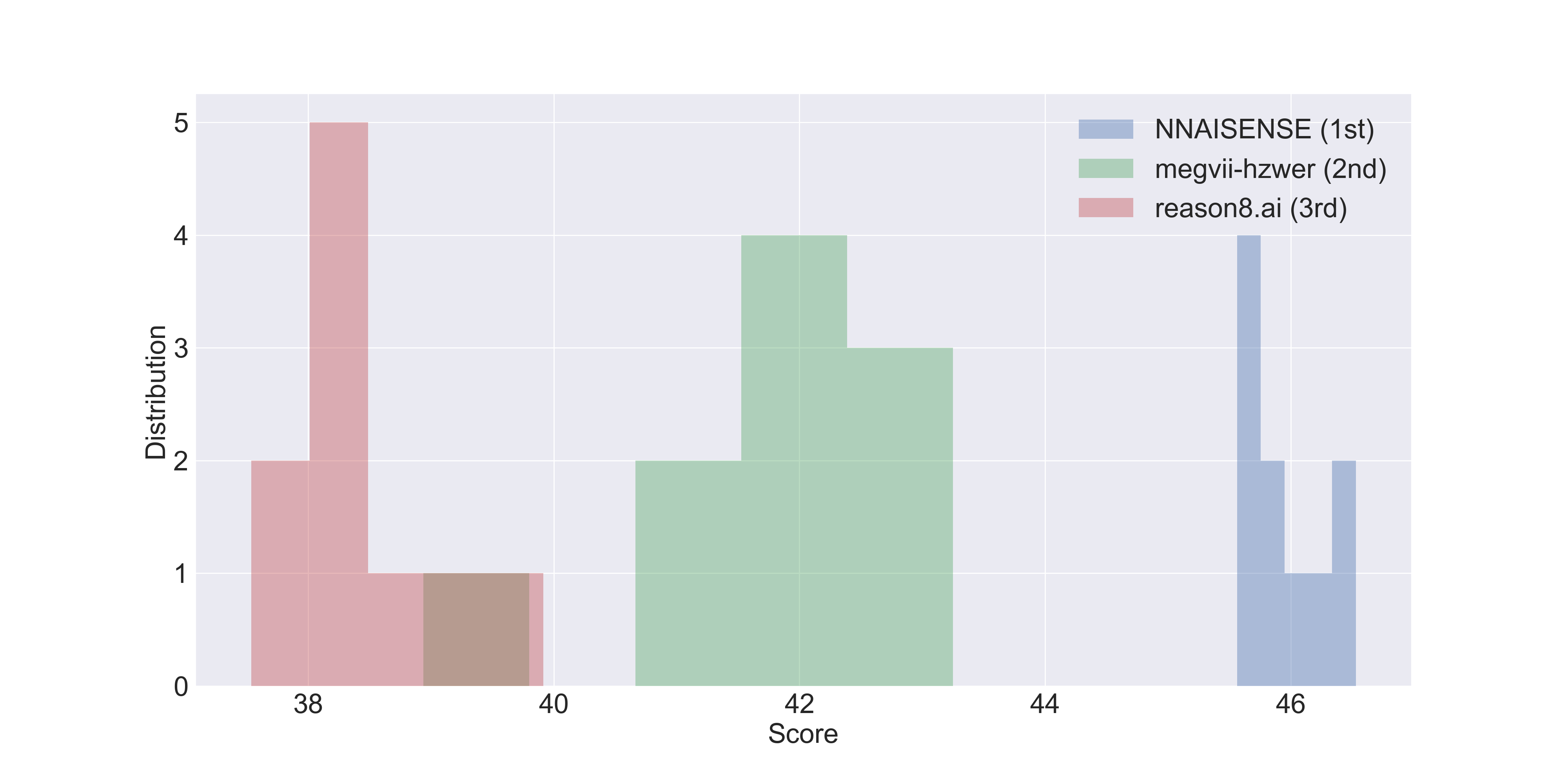}
\caption{Distribution of scores per simulation of the top 3 submitted entries in the Play-off Stage. In the final round of the challenge we ran 10 simulations in order to reduce the impact of randomness on the results. As it can be seen in the plot, the scores of top three participants were rather consistent between the simulations. This indicates that, despite stochasticity of simulations, our protocol allowed to determine the winner with high degree of confidence.}
\label{fig:top_3_distribution}
\end{figure}

The design of the Play-off stage allowed us to vary the number of simulations used to determine the winner. However, our initial choice of $10$ trials turned out to be sufficiently large to clearly determine the top places (Figure~\ref{fig:top_3_distribution}).

From the reports of top participants \cite{kidzinski2018l2rsolutions,jaskowski2018rltorunfast}, we observed that most of the teams (6 out of 9) used DDPG as the basis for their final algorithm while others used Proximal Policy Optimization (PPO) \cite{schulman2017proximal}. Similarly, in a survey we conducted after the challenge, from ten respondents (with mean scores $17.4$ and standard deviation $14.1$) five used DDPG, while two used PPO. This trend might be explained by the high computational cost of the environment, requiring the use of data efficient algorithms.

%In order to get better understanding of methods that participants used, we conducted a survey. From ten participants who responded, five used DDPG to train their final model. three used PPO and This, combined with the reports from our top participants  \cite{kidzinski2018l2rsolutions} give us an approximation of the methods tried. 

\section{Results}\label{s:results}
%\subsection{Analysis}
% I would include a discussion of figures five and six in the results section. This section could also use some sort of introduction/overview at the beginning
We conducted a post-hoc analysis of the submitted controllers by running the solutions on flat ground with no muscle weakness (i.e., \verb|max_obstacles=0, difficulty=0|). Only the last 5 seconds of each simulation were analyzed in order to focus on the cyclic phase of running. To compare simulation results with experimental data, we segmented the simulation into individual gait cycles, which are periods of time between when a foot first strikes the ground until the same foot strikes the ground again. For each submission, all gait cycles were averaged to generate a single representative gait cycle. We compared the simulations to experimental data of individuals running at 4.00 m/s \cite{hamner2013running} as it is the closest speed to the highest scoring submissions. 

% I would change the title of the first column to "Scores Over 40". You should also explain this in the figure legend. % changed to scores over 40
\begin{figure}[ht!]
  \centering
  \includegraphics[width=0.9\linewidth]{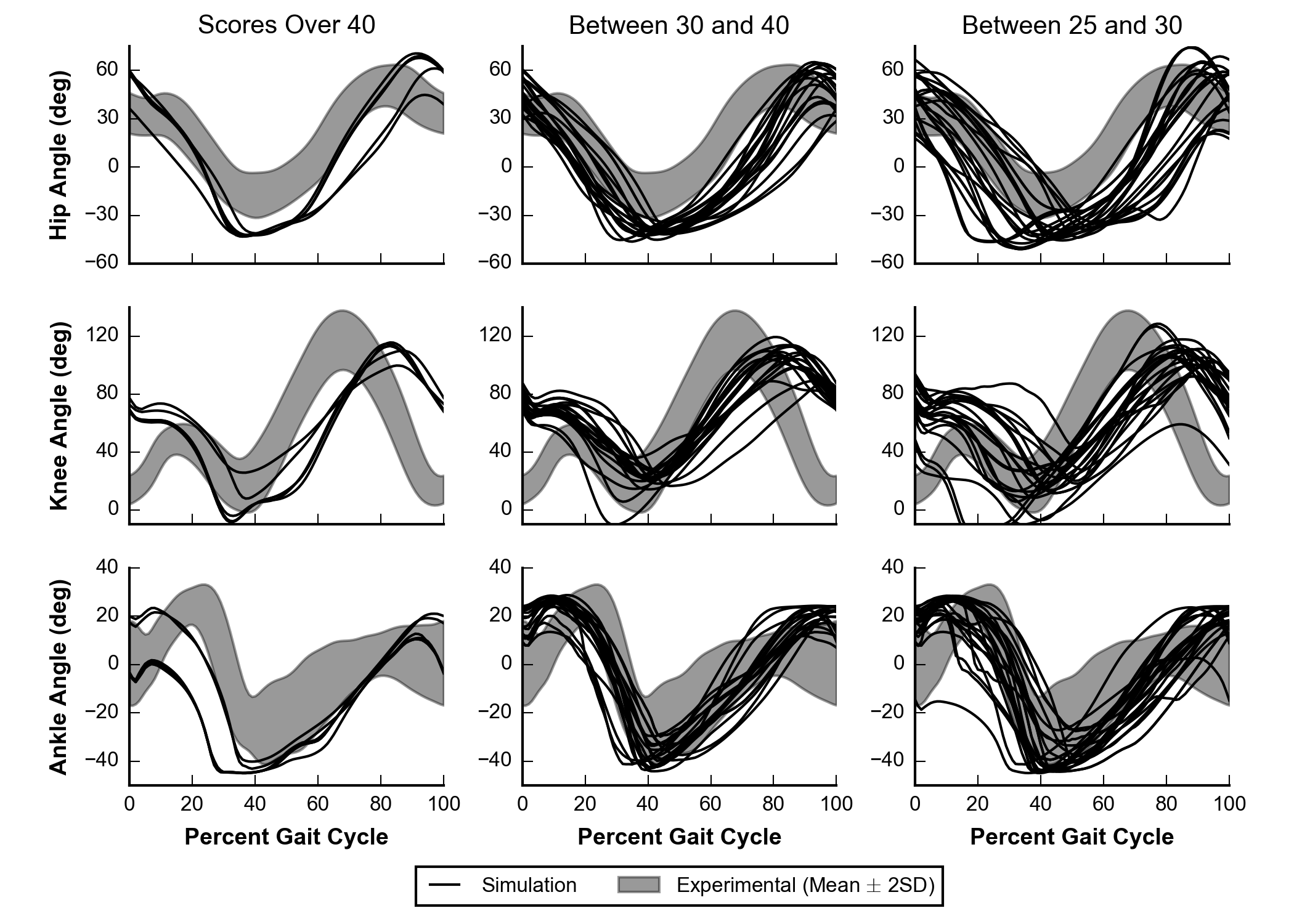}
  \caption{Simulated hip, knee, and ankle angle trajectories (black lines) compared to 20 experimental subjects running at 4.00 m/s (gray regions) \cite{hamner2013running}. Results are plotted from left to right with decreasing performance in three bins: scores over 40 (left), between 30 and 40 (middle), and between 25 and 30 (right). Positive values indicate flexion, and negative values indicate extension. 0\% gait cycle indicates when the foot first strikes the ground.}
  \label{fig:analysis}
\end{figure}

Figure \ref{fig:analysis} compares the simulated hip, knee, and ankle joint angle trajectories with experimental data, separated by three bins of scores: 1) over 40, 2) between 30 and 40, and 3) between 25 and 40. These bins represent solutions with the following rankings: 1) 1st through 5th, 2) 6th through 24th, and 3) 25th through 47th.
% The rationale for these bins is not completely clear. What is a high score? % added the corresponding rankings
Solutions in all of the score bins show some promising trends. For example, solutions in all three score bins have joints that are extending, shown by decreasing angle values, through the first 40\% of the gait cycle indicating that the models are pushing off the ground at this phase. Joint angles begin flexing, shown by increasing angle values during the last 60\% of the gait cycle in order to lift the leg up and avoid tripping.

% I changed the first sentence. The difference between ankle and hip/knee is not super obvious to me and I don't think important to the paragraph.
There were a few notable differences between the simulations and the running gait of humans. At the hip, simulations have a greater range of motion than experimental data as solutions both flex and extend more than is seen in human running. This could be due to the simple, planar model. At the knee, simulations had an excessively flexed knee at initial contact of the foot (i.e., 0\% of the gait cycle) and had a delayed timing of peak knee flexion compared to experimental data (i.e., around 80\% of the gait cycle compared to 65\% of the gait cycle).

Future work can be done to improve the current results. Improving the fidelity of the model could yield better results. For example, allowing the hip to adduct and abduct (i.e., swing toward and away from the midline of the body)  would allow the leg to clear the ground with less hip flexion and reduce the excessive hip range of motion. Testing different reward functions may also improve results, such as adding terms related to energy expenditure \cite{wang2012optimizing}. Finally, it is likely that the best solution still has not been found, and further improvements in reinforcement learning methods would help to search the solution space more quickly, efficiently and robustly.

% This seems like a rather abrupt ending. Could we move the impact section to the end? % added Discussion
\section{Discussion}

The impact of the challenge ranged across multiple domains. First, we stimulated new techniques in reinforcement learning. We also advanced and popularized an important class of reinforcement learning problems with a large set of output parameters (human muscles) and comparatively small dimensionality of the input (state of a dynamic system). Algorithms developed in the complex biomechanical environment also generalize to other reinforcement learning settings with highly-dimensional decisions, such as robotics, multivariate decision making (corporate decisions, drug quantities), stock exchange, etc.

This challenge also directly impacted the biomechanics and neuroscience communities. The control models trained could be extended to and validated in, for example, a clinical setting to help predict how a patient will walk after surgery \cite{ackermann2010optimality}. The controllers developed may also approximate human motor control and thus deepen our understanding of human movement. Moreover, by the analogy to Alpha Go, where reinforcement learning strategy outperforms humans \cite{silver2017mastering} due to broader exploration of the solution space, in certain human movements we may potentially find strategies more efficient in terms of energy or accuracy. Reinforcement learning is also a powerful tool for identifying deficiencies and errant assumptions made when building models, and so the challenge can improve on the current state-of-the-art for computational musculoskeletal modeling.

Our environment was setup using an open-source physics engine - a potential alternative for commercial closed-source MuJoCo, which is widely used in the reinforcement learning research community. Similarly, crowdAI.org--the platform on which the challenge was hosted--is also an open-source alternative to Kaggle\footnote{\url{https://kaggle.com/}}. By leveraging the agile infrastructure of crowdAI.org and components from OpenAI reinforcement learning environments\footnote{\url{https://github.com/kidzik/osim-rl-grader}}, we were able to seamlessly integrate the reinforcement learning setting (which, to the date, is not available in Kaggle).

This challenge was particularly relevant to the NIPS community as it brought together experts from both neuroscience and computer science. It attracted $442$ competitors with expertise in biomechanics, robotics, deep learning, reinforcement learning, computational neuroscience, or a combination. Several features of the competition ensured a large audience. Entries in the competition produced engaging (and sometimes comical) visuals of a humanoid moving through a complex environment. Further, we supplied participants with an environment that is easy to set-up and get started, without extensive knowledge of biomechanics.

\section{Affiliations and acknowledgments}
{\L}ukasz Kidzi\'nski, Carmichael Ong, Jennifer Hicks and Scott Delp are affiliated with Department of Bioengineering, Stanford University. Sharada Prasanna Mohanty, Sean Francis and Marcel Salathé are affiliated with Ecole Polytechnique Federale de Lausanne. Sergey Levine is affiliated with University of California, Berkeley.

The challenge was co-organized by the Mobilize Center, a National Institutes of Health Big Data to Knowledge (BD2K) Center of Excellence supported through Grant U54EB020405. It was partially sponsored by NVIDIA, Amazon Web Services, and Toyota Research Institute.

\appendix
\section{Appendix}\label{appendix}
\subsection{Installation}
\begin{figure}[ht!]
  \centering
  \includegraphics[width=0.4\linewidth]{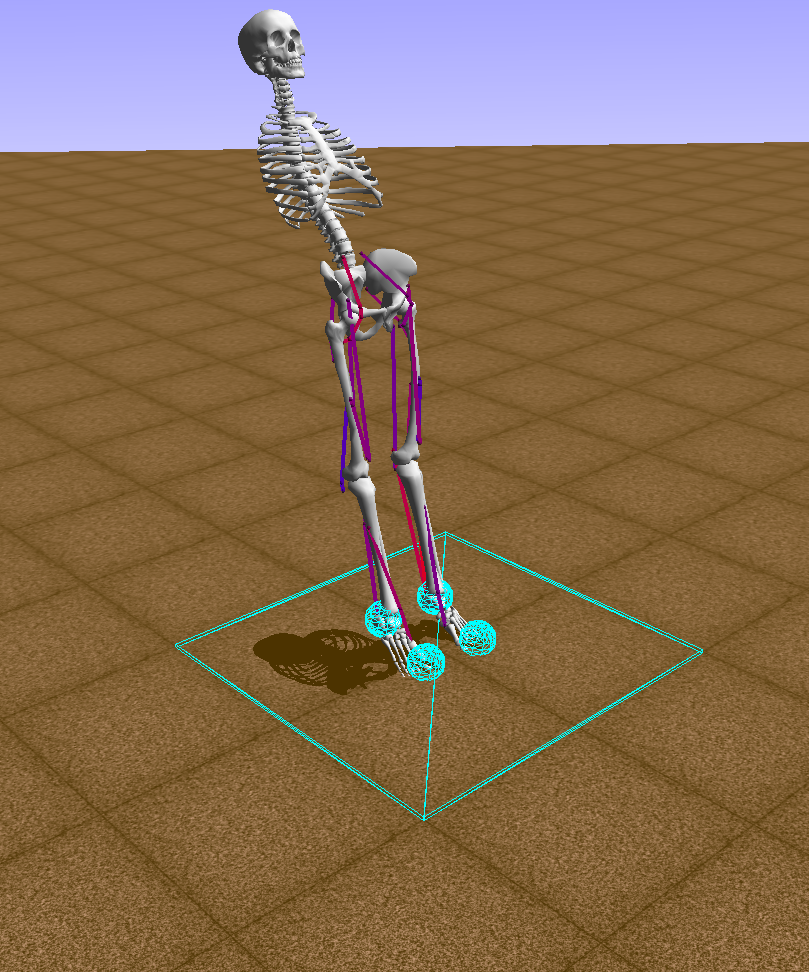}
  \caption{Visualization of the environment with random muscles activations after. This simulation is immediately visible to the user after following simple installation steps as described in Appendix.}
  \label{fig:training}
\end{figure}

We believe that the simplicity of use of the simulator (independently of the skills in computer science and biomechanics) contributed significantly to the success of the challenge. The whole installation process took around 1-5 minutes depending on the internet connection. To emphasize this simplicity let us illustrate the installation process.
Users were asked to install Anaconda (\url{https://www.continuum.io/downloads}) and then to install our reinforcement learning environment by typing
\begin{verbatim}
conda create -n opensim-rl -c kidzik opensim git
source activate opensim-rl
pip install git+https://github.com/kidzik/osim-rl.git
\end{verbatim}
Next, they were asked to start a python interpreter which allows interaction with the musculoskeletal model and visualization of the skeleton (Figure \ref{fig:training}) after running
\begin{verbatim}
from osim.env import GaitEnv
env = GaitEnv(visualize=True)
observation = env.reset()
for i in range(500):
    observation, reward, done, info = env.step(env.action_space.sample())
\end{verbatim}

\end{document}